\documentclass[10pt,twocolumn,letterpaper]{article}

\usepackage{cvpr}
\usepackage{times}
\usepackage{graphicx}
\usepackage{amsmath}
\usepackage{amssymb}

\usepackage[shortlabels]{enumitem}
\usepackage{amsmath,amssymb,amsfonts}
\usepackage{multirow}
\usepackage[cal=cm]{mathalfa}
\usepackage{subfigure}
\usepackage{afterpage}
\usepackage[algoruled,boxed,lined,ruled]{algorithm2e}
\usepackage{boxedminipage}
\usepackage{tikz}
\usepackage{pgfplots}
\pgfplotsset{compat=1.12}
\usepackage{float}
\usepackage{rotating}

\usepackage[pagebackref=true,breaklinks=true,letterpaper=true,colorlinks,bookmarks=false]{hyperref}

\cvprfinalcopy 

\def\cvprPaperID{5981} 

\ifcvprfinal\fi
\begin{document}

\title{SQWA: Stochastic Quantized Weight Averaging for Improving\\the Generalization Capability of Low-Precision Deep Neural Networks}

\author{Sungho Shin, Yoonho Boo, and Wonyong Sung\\
Department of Electrical and Computer Engineering\\
Seoul National University\\
Seoul, 08826 Korea\\
{\tt\small sungho.develop@gmail.com, wysung@snu.ac.kr}
}

\maketitle

\begin{abstract}
Designing a deep neural network (DNN) with good generalization capability is a complex process especially when the weights are severely quantized. Model averaging is a promising approach for achieving the good generalization capability of DNNs, especially when the loss surface for training contains many sharp minima. We present a new quantized neural network optimization approach, \textit{stochastic quantized weight averaging} (SQWA), to design low-precision DNNs with good generalization capability using model averaging. The proposed approach includes (1) floating-point model training, (2) direct quantization of weights, (3) capturing multiple low-precision models during retraining with cyclical learning rates, (4) averaging the captured models, and (5) re-quantizing the averaged model and fine-tuning it with low-learning rates. Additionally, we present a loss-visualization technique on the quantized weight domain to clearly elucidate the behavior of the proposed method. Visualization results indicate that a quantized DNN (QDNN) optimized with the proposed approach is located near the center of the flat minimum in the loss surface. With SQWA training, we achieved state-of-the-art results for 2-bit QDNNs on CIFAR-100 and ImageNet datasets. Although we only employed a uniform quantization scheme for the sake of implementation in VLSI or low-precision neural processing units, the performance achieved exceeded those of previous studies employing non-uniform quantization. 
\end{abstract}

\section{Introduction}
\label{sec5_intro}
Deep neural networks (DNNs) have demonstrated highly promising results in various applications~\cite{he2016deep,amodei2016deep,young2018recent,wu2016google}; however, they typically require a large number of parameters and considerable arithmetic costs. Weight quantization is considered the most practical model compression technique; in fact, DNNs do not require the precision of 32-bit floating-point operations, especially in inference. However, implementing a DNN using low-precision weights, such as one or two bits, is extremely challenging because the direct quantization of floating-point weights does not yield good results. Hence, various quantization and training methods have been developed~\cite{hwang2014fixed,courbariaux2015binaryconnect,fengfu2016ternary,zhou2017balanced,mishra2018apprentice,shin2019hlhlp,cai2019weight}.

The purpose of DNN training is to achieve good generalization capability. Thus, it may not be optimal to use quantized DNN (QDNN) designs that approximate floating-point weights using elaborate coding techniques. In recent years, loss surface visualization has helped to improve the generalization capability of DNNs~\cite{izmailov2018averaging,garipov2018loss,draxler2018essentially}. 
Fast geometric ensemble (FGE)~\cite{garipov2018loss} and stochastic weight averaging
(SWA)~\cite{izmailov2018averaging} have been proposed based on the observation that local minima attained by stochastic gradient descent (SGD) training are closely connected~\cite{garipov2018loss}. FGE and SWA capture multiple models during training and ensemble or average the models to obtain a well-generalized network. Model averaging moves the averaged model to the center of the loss surface especially when training with SGD causes sticking at the local minimum. 

In this study, we employed the model averaging technique to design a QDNN with improved generalization capability. We used cyclical learning rate scheduling for retraining of directly quantized network, and captured multiple low-precision models near the end of training. However, it is not straightforward to apply the previously developed SWA or FGE to QDNN design because the weight precision of the averaged model increases. For example, if we take the average of seven 2-bit models with ternary weights (-$\Delta$, 0, and +$\Delta$), then a 4-bit model is obtained(-7$\Delta$, -6$\Delta$, ..., 0, ..., +6$\Delta$, and +7$\Delta$). Thus, we must quantize it again to obtain a 2-bit model. Loss-surface aware DNN training is facilitated significantly by recently developed visualization techniques. However, the loss-surface of a QDNN is different from that of a floating-point model because the representation capability of a low-precision network is limited. In this study, we developed a new visualization technique for QDNNs by applying the quantization training algorithm. The new visualization method can successfully explain the mechanism of the proposed SQWA.

Our main contributions are as follows:
\begin{itemize}[nosep]
\item We presented a new QDNN training technique, SQWA, to improve the generalization capability of QDNNs. 
\item With the proposed SQWA training scheme, we achieved state-of-the-art results on CIFAR-100 and ImageNet datasets. 
\item We proposed a loss visualization method for low-precision quantized DNNs.
\end{itemize}

\section{Related works}
\label{sec5_related}
\subsection{Quantization of deep neural networks for efficient implementations}
\label{sec5_related_quantization}
Typically, the precision of parameters and data is reduced for efficient implementations in real-time signal processing system designs. While audio and video signal processing demands precision exceeding eight bits, many DNNs function well with lower precisions, such as one or two bits. Particularly, the performance of low-precision DNNs can be improved considerably by conducting retraining after quantization. Thus, quantization is a highly promising approach for the efficient implementation of DNNs. The quantization training algorithm, first proposed by~\cite{hwang2014fixed} and~\cite{courbariaux2015binaryconnect}, has been combined with various types of quantization methods such as symmetric uniform~\cite{fengfu2016ternary}, asymmetric uniform~\cite{zhou2017balanced}, non-uniform~\cite{miyashita2016convolutional}, and differentiable~\cite{choi2018pact,yang2019quantization,hou2016loss,hou2018loss} quantizers.
In recent years, a few elaborate techniques have been developed, such as employing knowledge distillation and carefully controlling the learning rate and bit-precision for improved generalization~\cite{mishra2018apprentice,polino2018model,shin2019hlhlp}. Weight normalization is adopted to avoid a long tail distribution of the model weights~\cite{cai2019weight}.

In this work, we focus on obtaining a good training scheme to optimize QDNNs. This approach is focused on developing well-generalized low-precision DNNs instead of developing elaborate quantization schemes. It is noteworthy that we only used the uniform quantization scheme for simplifying the hardware~\cite{umuroglu2017finn,ando2017brein} for inference. Non-uniform quantization can yield improved QDNN performance when the precision is the same; however, it demands additional operations, which can be time-consuming when hardware with conventional arithmetic blocks is involved.

\subsection{Stochastic weight averaging and loss-surface visualization}
\label{sec5_related_visual}
Stochastic gradient descent (SGD) is the most widely used method for DNN training. However, the loss surface for SGD contains many sharp minima~\cite{hochreiter1997flat}; thus, SGD-based training exhibits overfitting frequently. Many regularization techniques can be applied for alleviating this problem, such as L2-loss, dropout, and cyclical learning rate scheduling~\cite{van2017l2,srivastava2014dropout,smith2017cyclical}. The ensemble of models is known to increase the generalization capability. However, this method typically demands increased cost for training and inference. Fast geometric ensemble (FGE) is a recent technique for the ensemble of models~\cite{garipov2018loss}. Dropout~\cite{srivastava2014dropout} and dropconnect~\cite{wan2013regularization} can be interpreted as building an ensemble of models by weight averaging. 

SWA is a recently developed regularization technique; that is based on the weight averaging of models captured during training with cyclical learning rate scheduling~\cite{izmailov2018averaging}. SWA demonstrates excellent performances in many CNN models on various datasets. 

SWA can be explained using the loss visualization technique. The visualization method for representing three models in a single loss surface has been suggested in~\cite{garipov2018loss} and~\cite{izmailov2018averaging}. In those studies, training algorithms SWA and FGE were presented by discovering that the local minima trained with SGD were interconnected. Furthermore, because the loss surface for training and test were different, the minima found by SGD during training were not necessarily the best for the test data. Instead, the average of the models indicated a significantly improved generalization capability.~\figurename~\ref{sec5_fig_intuition} (a) depicts three models captured during cyclical learning rate scheduling and shows that the average is located near the center in the loss surface~\cite{izmailov2018averaging}.

SWA has been applied to low-precision training. Stochastic weight averaging in low-precision (SWALP) employs SWA for a cost-efficient training, where a low-precision (\eg, 8-bit) model is trained with cyclical learning rate scheduling and models are captured during training at the lowest learning rate in the cycle~\cite{yang2019swalp}. The captured models are then averaged to obtain the final full-precision model. Thus, SWALP is intended to design high-precision models and is vastly different from our work, which optimizes severely quantized models (\eg, 2-bit) for inference. 

\section{Quantization of DNN and loss surface visualization}
\label{sec5_geometric}
In this section, we explain how DNNs can be optimally quantized using a retraining method and then present a loss surface visualization method for QDNNs. To this end, we first revisit a previous method~\cite{izmailov2018averaging,garipov2018loss} to visualize three weight vectors in a single loss surface and explain the limitation when applied to QDNN loss surface visualization.

\begin{figure*}[t]
\centering
\subfigure[Conventional~\cite{izmailov2018averaging}]{\includegraphics[width=0.495\linewidth]{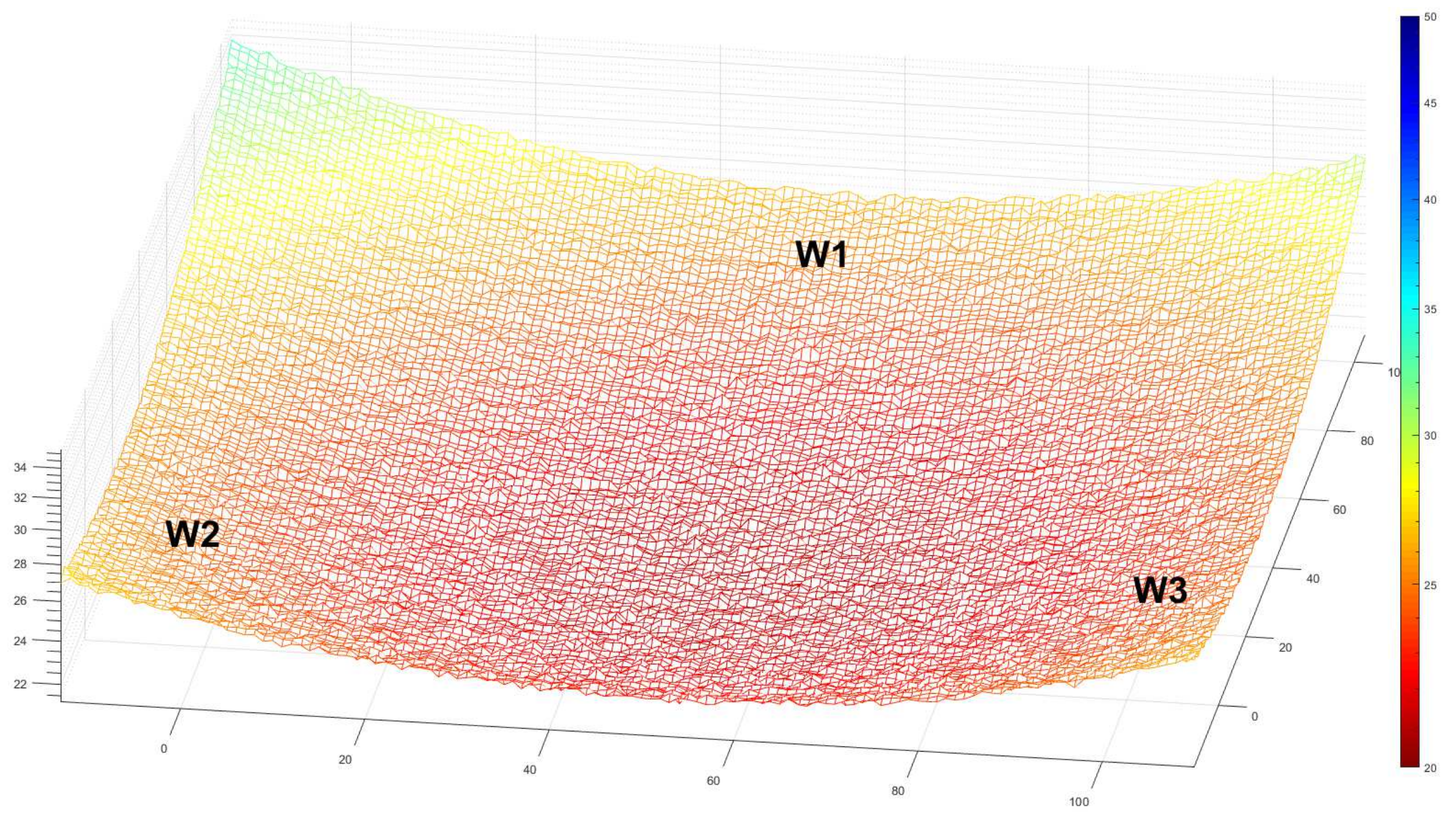}}
\subfigure[Ours]{\includegraphics[width=0.495\linewidth]{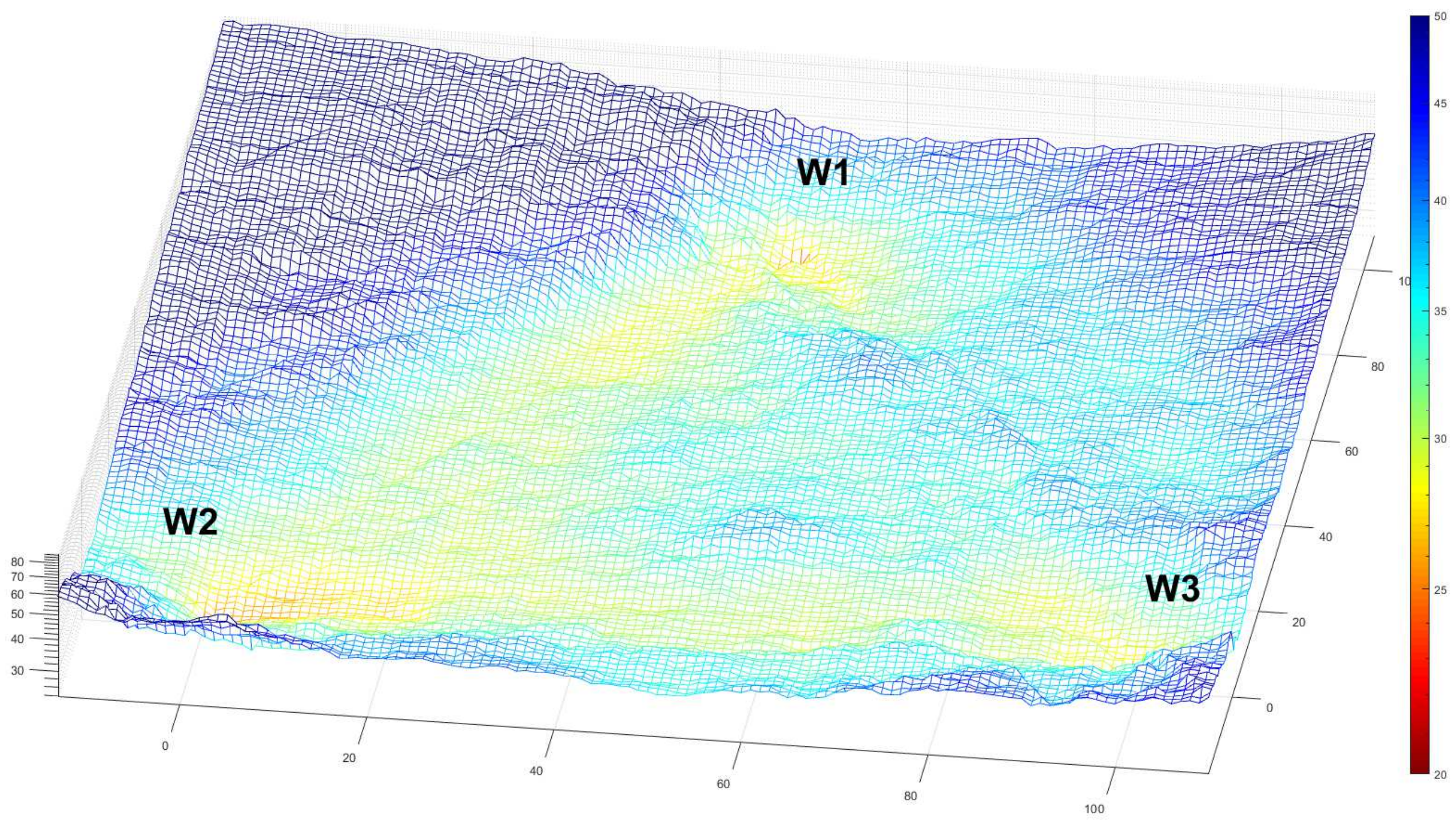}}
\caption{Visualization of three QDNNs in a single loss surface with the conventional method~\cite{izmailov2018averaging} (a) and ours (b). Three models are captured during fixed-point retraining. The points of $\mathbf{w1}$, $\mathbf{w2}$, and $\mathbf{w3}$ represent the captured models at 214th, 232th, and 250th epochs, respectively.}
\label{sec5_fig_general}  
\end{figure*}

\subsection{Quantization of deep neural networks}
\label{secsec5_quant}
The weight vector, $\mathbf{w}$, of a deep neural network can be quantized in $b$-bit using a symmetric uniform quantizer as follows:
\begin{align}
\text{Q}^{b}(\mathbf{w}) &= \text{sign}(\mathbf{w})\cdot\Delta\cdot\text{min}\Big\{\Big\lfloor\Big(\frac{|\mathbf{w}|}{\Delta}+0.5\Big)\Big\rfloor, \frac{(M-1)}{2}\Big\},\label{sec5_eq_quant}
\end{align}
where $\text{sign}(\cdot)$ is the sign function, $\Delta$ is the quantization step size, $M$ is the number of quantization levels that can be computed with $2^b-1$. A trained full-precision model can be directly quantized with Equation~\eqref{sec5_eq_quant}, but the performance will be significantly degraded when severe quantization such as 1- or 2-bit is employed. To relieve this problem, retraining on quantization domain is adopted in previous studies~\cite{hwang2014fixed,courbariaux2015binaryconnect,shin2017fixed,zhou2016dorefa} as follows:
\begin{gather}
l_{i} = \sum\limits_{j \in A_{i}}w_{ij}^{(q)}y_{j}^{(q)} \label{sec5_eq:logit} \\
y_{i}^{(q)}=\phi_{i}(l_{i}) \label{sec5_eq:forward} \\ 
\delta_{j}=\phi_{j}^{'}(l_{j})\sum\limits_{i \in P_{j}}\delta_{i}w_{ij}^{(q)}  \label{sec5_eq:backward}\\
\dfrac {\partial E}{\partial w_{ij}} = -\delta_{i}y_{j}^{(q)} \label{sec5_eq:gradient} \\
w_{ij, new} = w_{ij} - \eta \biggl\langle \dfrac {\partial E}{\partial w_{ij}} \biggr\rangle \label{sec5_eq:weight_update} \\
w_{ij, new}^{(q)} = Q_{ij}(w_{ij, new}) \label{sec5_eq:weight}
\end{gather}
where $l_{i}$ is the logit of the unit $i$, $\delta_{i}$ is the error signal of the unit $i$, $w_{ij}$ is the weight from the unit $j$ to the unit $i$, $y_{j}$ is the output activation of the unit $j$. $\eta$ is the learning rate, $A_{i}$ is the set of units anterior to the unit $i$, $P_{j}$ is the set of units posterior to the unit $j$, $Q(\cdot)$ is the weight quantizer, $\phi(\cdot)$ is the activation function. The superscript $(q)$ indicates the value is quantized, and $\langle\cdot\rangle$ is the average operation over the mini-batch. As described in Equation~\eqref{sec5_eq:logit} to ~\eqref{sec5_eq:weight}, the forward, backward, and gradient calculation is conducted with the quantized weights, but weight update adopts the full-precision parameters. This is because that the quantization step size, $\Delta$, is usually much larger than the computed gradients, $\dfrac {\partial E}{\partial \mathbf{w}}$. The weights are not changed if the gradient is directly updated to the quantized weights.

\subsection{Loss surface visualization for QDNNs}
\label{secsec5_visualization}
The visualization method in~\cite{izmailov2018averaging,garipov2018loss} shows the location of three weight vectors $\mathbf{w}_1$, $\mathbf{w}_2$, and $\mathbf{w}_3$. For locating these three models on the same loss surface, the projection vectors $\mathbf{u}$ and $\mathbf{v}$ are formed as follows:
\begin{gather}
\mathbf{u} = (\mathbf{w}_2 - \mathbf{w}_1) \label{sec5_eq_u} \\
\mathbf{v} = (\mathbf{w}_3 - \mathbf{w}_1) - \langle \mathbf{w}_3 - \mathbf{w}_1, \mathbf{w}_2 - \mathbf{w}_1 \rangle / \left\lVert \mathbf{w}_2 - \mathbf{w}_1\right\rVert^2  \label{sec5_eq_v} \\
\mathbf{\hat u} = \frac{\mathbf{u}}{\left\lVert \mathbf{u} \right\rVert} \label{sec5_eq_u_hat}\\
\mathbf{\hat v} = \frac{\mathbf{v}}{\left\lVert \mathbf{v} \right\rVert} \label{sec5_eq_v_hat}
\end{gather}
The normalized vectors $\mathbf{\hat u}$ and $\mathbf{\hat v}$ form an orthonormal basis in the plane containing $\mathbf{w}_1$, $\mathbf{w}_2$, and $\mathbf{w}_3$. These three vectors can be visualized on a Cartesian grid in the basis $\mathbf{\hat u}$ and $\mathbf{\hat v}$ using a set of points $P$.
\begin{align}
P = \mathbf{w}_1 + x\cdot\mathbf{\hat u} + y \cdot\mathbf{\hat v} \label{sec5_eq_p},
\end{align}
where $x$ and $y$ are the coordinates.

We trained the 2-bit ternary quantized ResNet-20~\cite{he2016deep} on the CIFAR-100 dataset~\cite{krizhevsky2009learning} using the retraining algorithm~\cite{hwang2014fixed}. After the performance has fully converged during the retraining process, we captured three quantized models $\mathbf{w}^{(q)}_1$, $\mathbf{w}^{(q)}_2$, and $\mathbf{w}^{(q)}_3$ with time intervals\footnote{$\mathbf{w}^{(q)}_1$, $\mathbf{w}^{(q)}_2$, and $\mathbf{w}^{(q)}_3$ that were captured at epochs 214, 232, and 250, respectively} on the training epoch.~\figurename~\ref{sec5_fig_general}~(a) visualizes the three quantized networks using Equations~\eqref{sec5_eq_u} to~\eqref{sec5_eq_p}. 


It is noted that $\mathbf{w}^{(q)}_1$, $\mathbf{w}^{(q)}_2$, and $\mathbf{w}^{(q)}_3$ are located at `$\mathbf{w1}$', `$\mathbf{w2}$', and `$\mathbf{w3}$', respectively. The limitation of this visualization method when applied to QDNNs is obvious. Even though $\mathbf{w}^{(q)}_1$, $\mathbf{w}^{(q)}_2$, and $\mathbf{w}^{(q)}_3$ are quantized weights, the other points between them are represented in full-precision. Thus, the exact shape of the loss surface cannot be determined when the weights are quantized. Hence, we plot the loss surface after quantizing the high-precision location vector, $P$. It is noteworthy that we can employ the full-precision weight vectors from Equation~\eqref{sec5_eq:weight_update} to compute Equation~\eqref{sec5_eq_u_hat} and~\eqref{sec5_eq_v_hat}. We denote these two normalized vectors as $\mathbf{\hat u}^{f}$ and $\mathbf{\hat v}^{f}$ to avoid confusion; subsequently, the three quantized vectors can be visualized on a Cartesian grid using a set of quantized points, $P^{q}$, for QDNNs as follows:
\begin{gather}
\mathbf{w}^{f} = \mathbf{w}^{f}_1 + x\cdot\mathbf{\hat u}^{f} + y \cdot\mathbf{\hat v}^{f}\label{sec5_eq_p_modify}\\
P^{q} = \text{sign}(\mathbf{w}^{f})\cdot\Delta\cdot\text{min}\Big\{\Big\lfloor\Big(\frac{|\mathbf{w}^{f}|}{\Delta}+0.5\Big)\Big\rfloor, \frac{(M-1)}{2}\Big\}\label{sec5_eq_quant_modify}
\end{gather}
where $\mathbf{w}^{f}_1$ is the full-precision weight vector that can be employed during retraining. We report the relationship of the three vectors $\mathbf{w}_1^{(q)}$, $\mathbf{w}_2^{(q)}$, and $\mathbf{w}_3^{(q)}$ obtained using the modified visualization method in~\figurename~\ref{sec5_fig_general} (b). The relationship of the three quantized weight vectors, which cannot be observed in~\figurename~\ref{sec5_fig_general} (a), is well represented. As they were captured in the epoch order (`$\mathbf{w1}$' $\rightarrow$ `$\mathbf{w2}$' $\rightarrow$ `$\mathbf{w3}$') during the retraining, a path along $\mathbf{w}_1^{(q)}$, $\mathbf{w}_2^{(q)}$ and $\mathbf{w}_3^{(q)}$ appeared. Because all of the points, $P^{(q)}$, were expressed in 2-bit quantized weights, the surface fluctuated strongly owing to quantization noise.

\section{SQWA algorithm}
\label{sec5_SQWA}
Training a DNN can be regarded as guiding a model to near the center of the loss-surface of the training data. The weight quantization of a DNN incurs a large perturbation to the model, and even a well-trained DNN exhibits poor performance after a severe quantization. Thus, retraining is typically employed to return a model to the center of the training loss surface. The conventional fine-tuning approach that employs a low learning rate seeks to obtain a permissible nearby minimum in the quantized domain. In our opinion, this can be improved by employing more aggressive training methods. 

The proposed SQWA retrains the quantized model using cyclical learning rate scheduling instead of low learning rates for fine-tuning. We captured multiple models during retraining and obtained the average of the captured models. It is noteworthy that the averaging process increases the bit-precision of the model. For example, if we take the average of seven ternary models, then a 4-bit model is obtained. Thus, we must re-quantize the averaged model, followed by fine-tuning using low learning rates. 

\begin{figure}[t]
\centering
\subfigure[SWA]{\includegraphics[width=.85\linewidth]{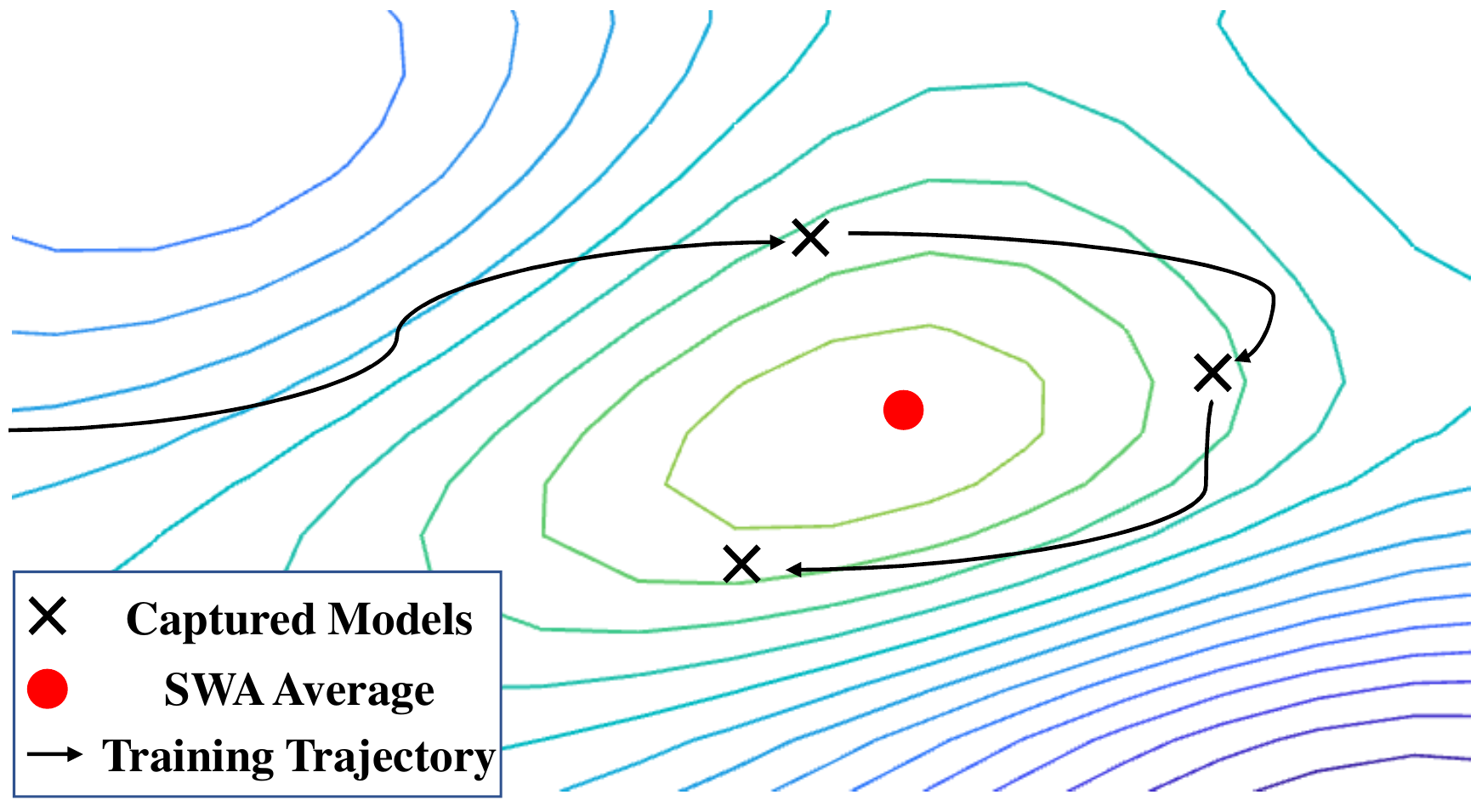}}
\subfigure[SQWA (ours)]{\includegraphics[width=.85\linewidth]{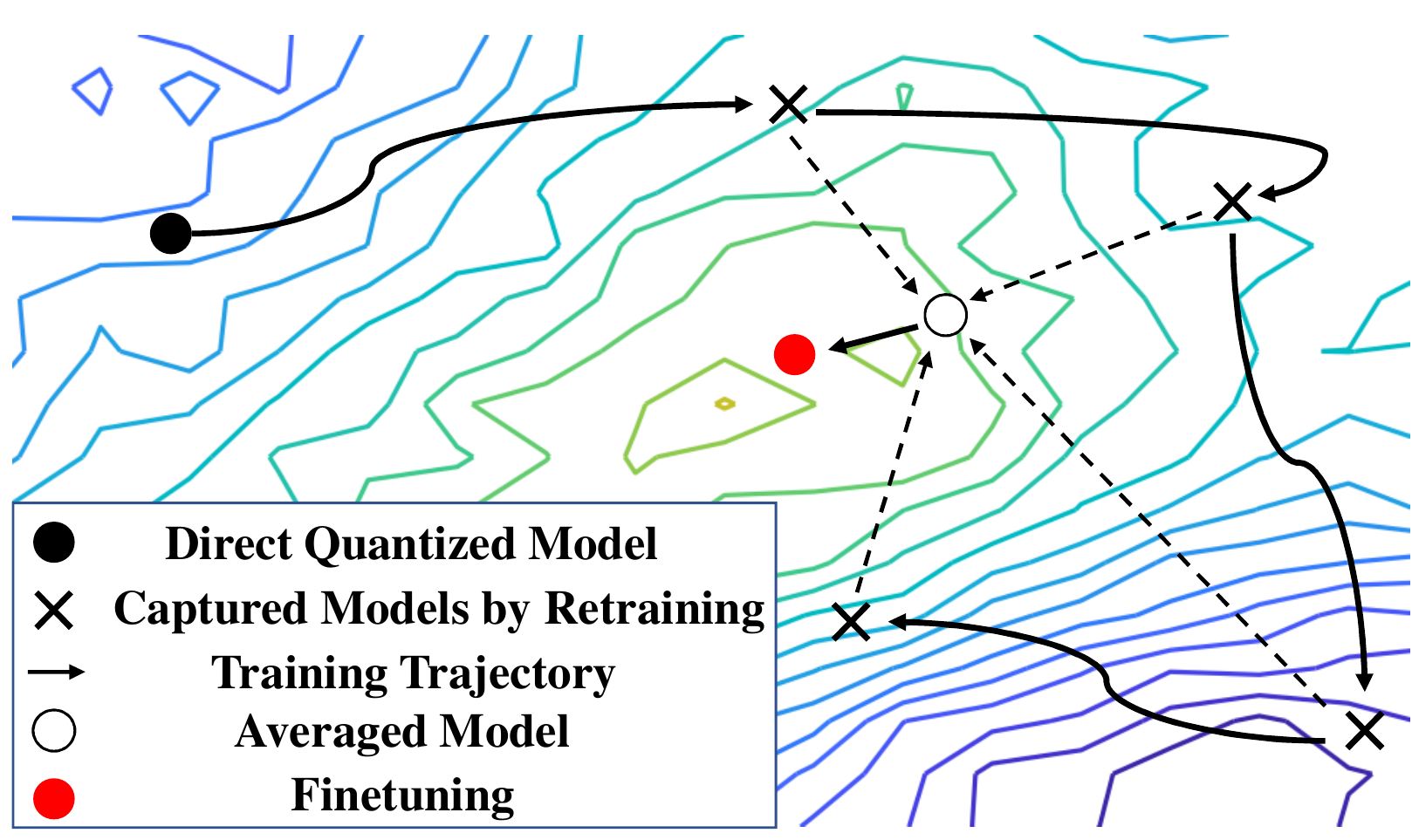}}
\caption{Intuitions of the SWA and the SQWA.}
\label{sec5_fig_intuition}  
\end{figure}

SQWA can be explained as shown in~\figurename~\ref{sec5_fig_intuition}. The difference is that optimization using the quantized loss surface is required. As shown in Section~\ref{sec5_geometric}, the quantized loss surface is rough when compared with its high-precision counterpart. Thus, optimization is more difficult with low learning rates. Cyclical learning rate scheduling, which uses high and low learning rates alternately, and weight averaging are more effective than fine-tuning for traversing the rugged loss surface. 

We demonstrate the entire workflow of the SQWA in~\figurename~\ref{sec5_fig_intuition} (b). The details are provided as follows.

\noindent\textbf{Pretrain a full-precision model:} 
\noindent We used high-performance floating-point models for the design of the QDNN, instead of directly designing a QDNN from scratch. This approach is more convenient considering the GPU-dominant training facilities available. According to our experiments, the performance of a quantized model is closely related to that of the original floating-point network. Thus, good training methods such as knowledge distillation (KD)~\cite{hinton2015distilling} or SWA~\cite{izmailov2018averaging} are necessitated.

\noindent\textbf{Quantize the full-precision model and retraining with cyclical learning rate scheduling:}
\noindent We first quantized the full-precision model from step 1 and then conducted retraining on the quantization domain with cyclical learning rate scheduling. We adopted discrete cyclical learning rate scheduling for a better generalization~\cite{jastrzkebski2017three}. Detailed guidelines for scheduling are as follows. First, we define all values of the learning rates for the full-precision model as $\eta_{f}$. Then, the maximum and minimum values of the cyclic learning rate scheduling are determined as $\eta_{\text{cycleMax}} = \frac{\text{max}(\eta_{f})}{10}$ and $\eta_{\text{cycleMin}} = \frac{\text{min}(\eta_{f})}{10}$, respectively. These values of the learning rate are highly related to the quantization error. The quantized weights, $\mathbf{w}^{(q)}$, can be interpreted as adding a quantization noise, $\mathbf{n}$, to the full-precision weight $\mathbf{w}^{(f)}$. The quantization noise $\mathbf{n}$ increases as the number of quantization bit, $b$, decreases. It is noteworthy that performing a direct quantization with low-precision, such as one or two bits, typically degrades the performance significantly. Thus, the smaller the number of bits, the larger is the required learning rate for recovering the performance. Because our SQWA training method is designed for severe quantizations (\ie, a 2-bit ternary model), $\frac{\text{max}(\eta_{f})}{10}$ would be a good choice.

One period of the discrete cyclical learning rate, $c$, is a hyperparameter that affects the training performance. The appropriate value of $c$ is four to six epochs in our experiments. Thus, one or two learning rate steps can be considered between $\eta_{\text{cycleMax}}$ and $\eta_{\text{cycleMin}}$ to form discrete cyclical learning rate scheduling. We captured the models during training at the lowest learning rate (\ie, $\eta_{\text{cycleMin}}$).

\noindent\textbf{Averaging the captured models:}
\noindent The third step is averaging the captured low-precision models. Model averaging improves the generalization capability by moving the averaged model to the middle of the loss surface~\cite{izmailov2018averaging}. The number of captured models for averaging affects SQWA training. When employing a 2-bit ternary symmetric uniform quantizer, for example, each captured weight is represented as $-\Delta$, 0, and $\Delta$. If we select seven captured weights for averaging, the averaged model has the representation level of $-7\Delta$, $-6\Delta$, ..., 0, ...,  $6\Delta$, and $7\Delta$, which is a 4-bit QDNN. Averaging too few models will degrade the final performance, whereas averaging too many networks will render the training less efficient.

\noindent\textbf{Re-quantization and fine-tuning of the averaged model:}
\noindent The final goal of SQWA is to obtain a low-precision model, such as a 2-bit model; thus, we must quantize the averaged model into a low-precision one and fine-tune it with relatively low learning rates. We employed a monotonically decreasing learning rate scheduling for this step. Thus, we adopted the initial learning rate of 0.1$\eta_{\text{cycleMax}}$ and trained three or four epochs. It is noteworthy that we decreased the learning rate at every epoch.

More detailed information and experimental results of our proposed method are reported in Section~\ref{sec5_experimental}.

\section{Experimental results}
\label{sec5_experimental}
We evaluate the proposed SQWA method using the CIFAR-100~\cite{krizhevsky2009learning} and ImageNet~\cite{russakovsky2015imagenet} datasets. 

\begin{table}[t]
\centering
\caption{Train and test accuracies (\%) of the full-precision model candidates for SQWA training. `Conventional' means training without special techniques, `KD' represents knowledge distillation, and `SWA' is stochastic weight averaging.}
\label{table:SQWA_fullprecision}
\begin{tabular}{ccc}
\hline\hline
             & Train Acc. & Test Acc. \\ \hline
Conventional & 90.12      & 68.43     \\ 
KD~\cite{hinton2015distilling} & 87.55      & 71.06     \\ 
SWA~\cite{izmailov2018averaging}          & 90.44      & 70.45     \\ 
KD + SWA      & 87.02      & \textbf{71.26}     \\ \hline\hline
\end{tabular}
\end{table}

\begin{table*}[t]
\centering
\caption{Train and test accuracies (\%) of the quantized model during retraining with cyclical learning rate scheduling on CIFAR-100 dataset. The left column represents the result obtained at the beginning phase of retraining, while the right shows that at the last phase, 214th to 250th epochs. `Avg.' means the averaged model using 7 models during cyclical learning rate training with specific epochs, `Direct' represents the direct quantization results of the averaged model, and `Fine-tune' is the result after fine-tuning of direct quantized network.}
\label{table:SQWA_averaged}
\begin{tabular}{ccc|ccc}
\hline\hline
Epoch (precision)                & Train Acc. & Test Acc. & Epoch (precision)                 & Train Acc.     & Test Acc.      \\ \hline
76 (2-bit)                  & 72.80      & 64.56     & 250 (2-bit)                 & 76.12          & 66.13          \\ 
70 (2-bit)                  & 73.28      & 65.20     & 244 (2-bit)                 & 75.68          & 66.38          \\ 
64 (2-bit)                  & 72.32      & 64.03     & 238 (2-bit)                 & 75.33          & 66.33          \\ 
58 (2-bit)                  & 73.16      & 65.14     & 232 (2-bit)                 & 75.32          & 65.96          \\ 
52 (2-bit)                  & 72.03      & 64.43     & 226 (2-bit)                 & 75.14          & 65.65          \\ 
46 (2-bit)                  & 72.00      & 64.54     & 220 (2-bit)                 & 75.08          & 66.02          \\ 
40 (2-bit)                  & 72.27      & 64.60     & 214 (2-bit)                 & 75.78          & 66.41          \\ \hline\hline
Avg. (4-bit)          & 76.53      & 67.94     & Avg. (4-bit)          & 78.95          & 68.81          \\ 
Direct (2-bit) & 62.89      & 56.93     & Direct (2-bit) & 70.31          & 62.52          \\ 
Fine-tune (2-bit)     & 74.25      & 66.75     & Fine-tune (2-bit)     & \textbf{76.83} & \textbf{67.75} \\ \hline\hline
\end{tabular}
\end{table*}

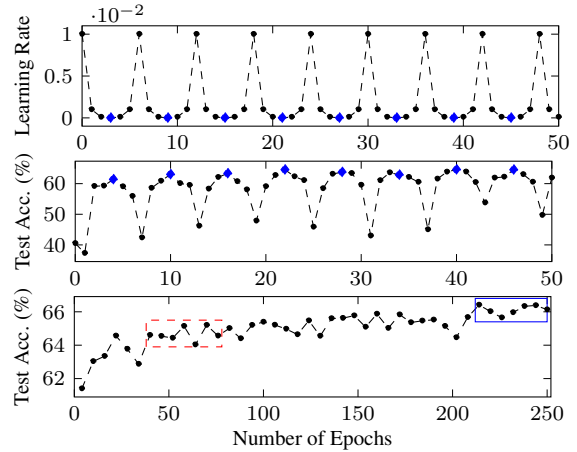
\begin{figure}[t]
\centering
\tikzset{mark options={mark size=1}}
	
\subfigure{
\begin{tikzpicture}
    \begin{axis}[
	width=0.95\linewidth,
	height = 0.35\linewidth,
	compat=1.12,
	xmin=0,
	xmax=50,
	log basis x={10}, 
	xtick pos=both, 
	xtick align=inside, 
	major tick style={line width=0.010cm, black},
	 major tick length=0.10cm,
        ylabel=Learning Rate,
	xlabel shift=-3pt,
	ylabel shift=-3pt,
	scatter/classes={
	a={black},
    b={black},
    c={black},
    d={blue,mark=diamond*,mark size=2},
    e={black},
    f={black}},
    tick label style={font=\footnotesize},
    label style={font=\footnotesize}
    ]]
	\addplot[scatter,scatter src=explicit symbolic,color=black,dashed] file{cycle_learning_rate_cifar100.txt}; 
    \end{axis}
   \end{tikzpicture}}
    \vskip -0.4cm
   \subfigure{\begin{tikzpicture}
    \begin{axis}[
	width=0.95\linewidth,
	height = 0.35\linewidth,
	compat=1.12,
	xmin=0,
	xmax=50,
	log basis x={10}, 
	xtick pos=both, 
	xtick align=inside, 
	major tick style={line width=0.010cm, black},
	 major tick length=0.10cm,
        ylabel=Test Acc. (\%),
	xlabel shift=-3pt,
	ylabel shift=-3pt,
	scatter/classes={
	a={black},
    b={black},
    c={black},
    d={blue,mark=diamond*,mark size=2},
    e={black},
    f={black}},
    tick label style={font=\footnotesize},
    label style={font=\footnotesize}
    ]]
	\addplot[scatter,scatter src=explicit symbolic,color=black,dashed] file{test_curve_cifar100_SWA_parser_test.txt}; 	
    \end{axis}
   \end{tikzpicture}}
    \vskip -0.4cm
      \subfigure{\begin{tikzpicture}
    \begin{axis}[
	width=0.95\linewidth,
	height = 0.35\linewidth,
	compat=1.12,
	xmin=0,
	xmax=252,
	log basis x={10}, 
	xtick pos=both, 
	xtick align=inside, 
	major tick style={line width=0.010cm, black},
	 major tick length=0.10cm,
        xlabel=Number of Epochs,
        ylabel=Test Acc. (\%),
	xlabel shift=-3pt,
	ylabel shift=-3pt,
	scatter/classes={
    a={black},
    b={black},
    c={black},
    d={black},
    e={black},
    f={black}},
    tick label style={font=\footnotesize},
    label style={font=\footnotesize}
    ]]
	\addplot[scatter,scatter src=explicit symbolic,color=black,dashed] file{sampled_cifar100_test_acc.txt}; 
	\draw[color=red,dashed] (38,63.9) rectangle (78,65.5);
	\draw[color=blue,solid] (212,65.4) rectangle (250,66.8);
    \end{axis}
   \end{tikzpicture}}

\caption{(\textbf{Top}): Cyclical learning rate scheduling for CIFAR-100 dataset, (\textbf{Middle}): the test accuracy curve with ResNet20, (\textbf{Bottom}): the sampled test accuracy curve from the every minimum learning rates with ResNet20.}
\label{sec5_fig_lrschedule_cifar100}
\end{figure}

\subsection{CIFAR-100}
\label{secsec5_cifar100}
\textbf{Network and hyperparameter configuration:} We trained ResNet-20~\cite{he2016deep} and MobileNetV2~\cite{sandler2018mobilenetv2} for the CIFAR-100 dataset. The training hyperparameters are as follows. For full-precision training, the batch size was 128 and the number of epochs trained was 175. An SGD optimizer with a momentum of 0.9 was used. The learning rate began at 0.1 and decreased by 0.1 times at the 75th and 125th epochs. Additionally, L2-loss was added with a scale of 5e-4. For QDNN retraining, the batch size and optimizer were the same as those of the full-precision one. The cyclical learning rate scheduling for retraining is described in~\figurename~\ref{sec5_fig_lrschedule_cifar100} (Top).

We captured the quantized models at the minimum points of the cyclical learning rate scheduling and obtained their average. To fine-tune the averaged model, the initial learning rate was set as 0.001 and decreased by 0.1 times at every epoch. We only performed three to four epochs for the fine-tuning. We did not employ L2-loss for the QDNN training as it conflicted with the clipping of the quantization.

\noindent\textbf{Results:} As described in Section~\ref{sec5_SQWA}, SQWA requires a pretrained full-precision model. We compare the models developed with KD~\cite{hinton2015distilling} and SWA~\cite{izmailov2018averaging} in~\tablename~\ref{table:SQWA_fullprecision}. The best full-precision model was trained by applying both KD and SWA. Its test accuracy was 71.26\%. We selected this network as the original full-precision model.

In the next step of the SQWA training process, we performed QDNN training with cyclical learning rate scheduling, as depicted in~\figurename~\ref{sec5_fig_lrschedule_cifar100} (Top). We captured the models when the learning rates were the lowest in the cycles (\ie, blue diamonds in~\figurename~\ref{sec5_fig_lrschedule_cifar100} (Top and Middle)). To select the models for averaging, we considered two groups of the networks, as depicted in~\figurename~\ref{sec5_fig_lrschedule_cifar100} (Bottom). The first group (dashed  red box) was selected at the beginning of the training and the other group (solid blue box) was captured after a sufficient number of training epochs has elapsed. We employed seven models for both groups, and their performances are compared in~\tablename~\ref{table:SQWA_averaged}.

More specifically, the models in the first group were captured between the 40th and 76th\footnote{The training performance was too low to capture for models earlier than 40th epochs.} epochs. Their test accuracies were approximately 64.7\% and the averaged model demonstrated an accuracy of 67.94\%. It is noteworthy that we took the average of seven 2-bit ternary QDNNs ($-\Delta$, 0, and $\Delta$), of which the resultant model was a 4-bit ($-7\hat\Delta$, $-6\hat\Delta$, ..., 0, ..., $6\hat\Delta$, and $7\hat\Delta$) QDNN. We conducted re-quantization and fine-tuned the averaged model to obtain a final 2-bit QDNN, which yielded a test accuracy of 66.75\%. The result of the second group was significantly better than that of the first group. The averaged 4-bit model yielded a test accuracy of 68.81\%. After the fine-tuning, the final performance of the 2-bit QDNN was 67.75\%. 
From these results, we can deduce the following:
\begin{itemize}[nosep]
    \item SQWA can be fully utilized when the models are captured after a sufficient convergence.
    \item Based on the observation of the direct quantization results in~\tablename~\ref{table:SQWA_averaged}, the second group forms a wider minima in the loss surface than the first group. Because direct quantization can be interpreted as a noise injection operation, less performance degradation suggests that the model is laying in a wider minimum or at the center of the loss surface.
\end{itemize}

\begin{table}[t]
\centering
\caption{Comparison with literature in terms of the test accuracy (\%) for quantized ResNet20 and MobileNetV2 on CIFAR-100.}
\label{table:SQWA_compare_cifar100}
\setlength\tabcolsep{4pt}
\begin{tabular}{ccc}
\hline\hline
ResNet20    & Quant Level  & Test Acc. \\ \hline
DoReFa-Net~\cite{zhou2016dorefa}& 4-level & 66.95          \\ 
Residual~\cite{guo2017network}&  4-level  & 65.97          \\ 
LQ-Net~\cite{zhang2018lq} &    4-level & 66.53          \\ 
WNQ~\cite{cai2019weight}   &    4-level  & 67.42          \\ 
HLHLp~\cite{shin2019hlhlp}  &   Ternary  & 66.44          \\ 
KDQ~\cite{shin2019empirical} &   Ternary   & 67.00          \\ 
SQWA (ours)& Ternary & \textbf{67.75}          \\ \hline
KDQ~\cite{shin2019empirical} &   Binary   & 60.14          \\ 
SQWA (ours)& Binary & \textbf{62.32}     \\ \hline\hline
MobileNetV2    & Quantization Level  & Test Acc. (\%) \\ \hline
L2Quant~\cite{anwar2015fixed}& Ternary & 74.97          \\ 
HLHLp~\cite{shin2019hlhlp}&  Ternary  & 75.51          \\ 
SQWA (ours) &    Ternary & \textbf{76.73}           \\ \hline\hline
\end{tabular}
\end{table}

\begin{figure*}[t]
    \centering
    \subfigure[Train (\cite{izmailov2018averaging})]{\includegraphics[width=0.48\linewidth]{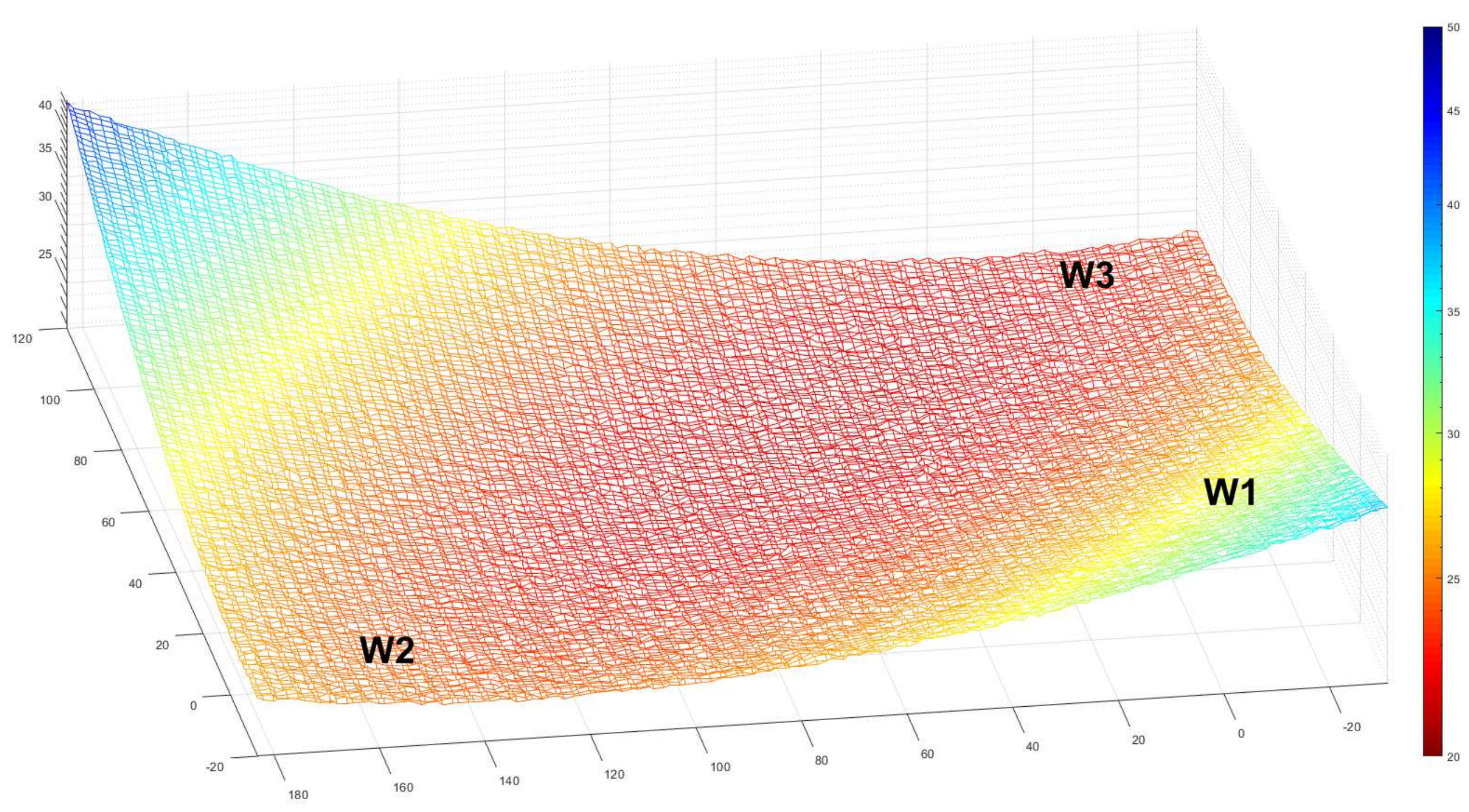}}\hfill
    \subfigure[Train (ours)]{\includegraphics[width=0.48\linewidth]{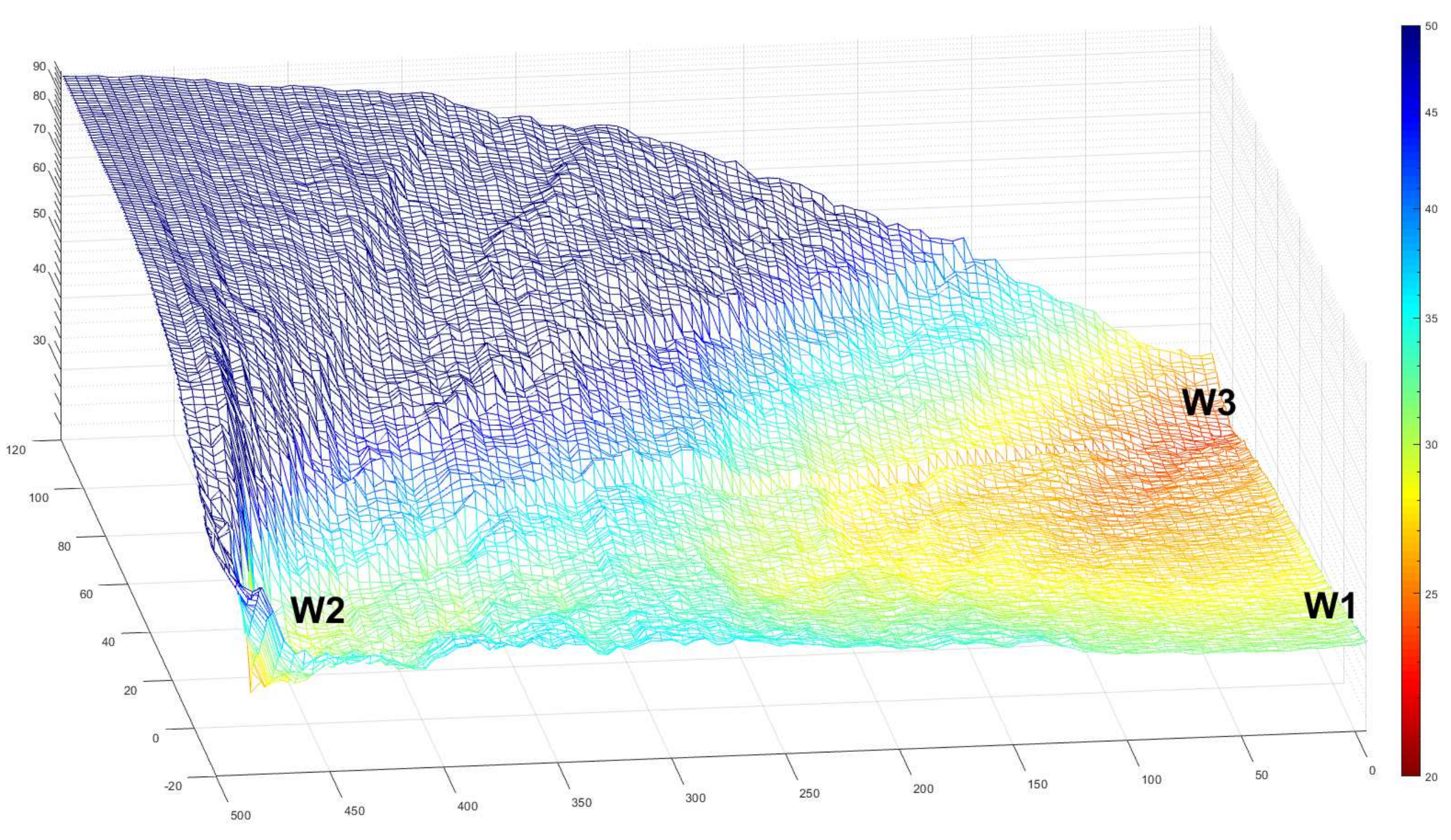}}
    \caption{Visualization in terms of train accuracies of three quantized models on a single loss surface. (a) is depicted by~\cite{izmailov2018averaging} and (b) is by ours. The points of `$\mathbf{w2}$', `$\mathbf{w1}$', and `$\mathbf{w3}$' represent `Epoch 214', `Direct', and `Fine-tune' in~\tablename~\ref{table:SQWA_averaged}, respectively.}
    \label{fig_compare_visual_cifar100}
\end{figure*}

We compare our SQWA results with those of previous studies in~\tablename~\ref{table:SQWA_compare_cifar100}. Our proposed SQWA method outperforms the methods of previous studies. In particular, SQWA indicated 0.8\%, 1.78\%, 1.22\%, and 0.33\% higher test accuracies than DoReFa-Net~\cite{zhou2016dorefa}, Residual~\cite{guo2017network}, LQ-Net~\cite{zhang2018lq}, and WNQ~\cite{cai2019weight}, respectively. This result is encouraging as the previous studies employed 2-bit 4-level quantizers, whereas we adopted 2-bit ternary and 1-bit binary quantizer. Furthermore, we compare our result to those involving 2-bit ternary and 1-bit binary quantizer. Our method with 2-bit ternary quantizer outperformed HLHLp~\cite{shin2019hlhlp} and KDQ~\cite{shin2019empirical} in terms of test accuracy by 1.31\% and 0.75\%, respectively. For the binary weights, SQWA achieves 2.18\% higher accuracy than KDQ. It should be noted that KDQ improves the performance of the QDNN using the KD. Additionally, we exploit the KD technique to obtain a high-performance full-precision model. Because SQWA outperforms KDQ, it suggests that SQWA training methods can combine with KD. Furthermore, we evaluated the proposed SQWA method using MobileNetV2, which has a larger number of parameters than ResNet20. We trained a full-precision MobileNetV2 with KD and SWA and achieved a test accuracy of 77.64\%. We exploited SQWA with the same cyclical learning rate scheduling used in the ResNet20 experiment. After a sufficient number of epochs, we captured seven models to establish a 4-bit averaged model and fine-tuned it. Our final 2-bit MobileNetV2 yielded the test accuracy that was 1.76\% and 1.22\% higher than those of L2Quant~\cite{anwar2015fixed} and HLHLp~\cite{shin2019hlhlp}, respectively.

\begin{figure}[t]
\centering
\tikzset{mark options={mark size=1}}
\subfigure{
\begin{tikzpicture}
    \begin{axis}[
	width=0.95\linewidth,
	height = 0.35\linewidth,
	compat=1.12,
	xmin=0,
	xmax=50,
	log basis x={10}, 
	xtick pos=both, 
	xtick align=inside, 
	major tick style={line width=0.010cm, black},
	 major tick length=0.10cm,
        ylabel=Learning Rate,
	xlabel shift=-3pt,
	ylabel shift=-3pt,
	scatter/classes={
	a={black},
    b={black},
    c={black},
    d={blue,mark=diamond*,mark size=2},
    e={black},
    f={black}},
    tick label style={font=\footnotesize},
    label style={font=\footnotesize}
    ]]
	\addplot[scatter,scatter src=explicit symbolic,color=black,dashed] file{cycle_learning_rate_imagenet.txt}; 
    \end{axis}
   \end{tikzpicture}}
   \vskip -0.4cm
   \subfigure{\begin{tikzpicture}
    \begin{axis}[
	width=0.95\linewidth,
	height = 0.35\linewidth,
	compat=1.12,
	xmin=0,
	xmax=50,
	log basis x={10}, 
	xtick pos=both, 
	xtick align=inside, 
	major tick style={line width=0.010cm, black},
	 major tick length=0.10cm,
        ylabel=Top-1 Acc. (\%),
	xlabel shift=-3pt,
	ylabel shift=-3pt,
	scatter/classes={
	a={black},
    b={black},
    c={black},
    d1={blue,mark=diamond*,mark size=2},
    e={black},
    f={black}},
    tick label style={font=\footnotesize},
    label style={font=\footnotesize}
    ]]
	\addplot[scatter,scatter src=explicit symbolic,color=black,dashed] file{imagenet_resnet18_top1.txt}; 
    \end{axis}
   \end{tikzpicture}}
   \vskip -0.4cm
    \subfigure{\begin{tikzpicture}
    \begin{axis}[
	width=0.95\linewidth,
	height = 0.35\linewidth,
	compat=1.12,
	xmin=0,
	xmax=252,
	log basis x={10}, 
	xtick pos=both, 
	xtick align=inside, 
	major tick style={line width=0.010cm, black},
	 major tick length=0.10cm,
        xlabel=Number of Epochs,
        ylabel=Top-1 Acc. (\%),
	xlabel shift=-3pt,
	ylabel shift=-3pt,
	scatter/classes={
    a={black},
    b={black},
    c={black},
    d={black},
    e={black},
    f={black}},
    tick label style={font=\footnotesize},
    label style={font=\footnotesize}
    ]]
	\addplot[scatter,scatter src=explicit symbolic,color=black,dashed] file{imagenet_resnet18_top1_lowlr.txt}; 
	\draw[color=red,dashed] (200,66.9) rectangle (240,68.6);
    \end{axis}
   \end{tikzpicture}}
\caption{(\textbf{Top}): Cyclical learning rate scheduling for ImageNet dataset, (\textbf{Middle}): a validation top-1 accuracy curve with ResNet18, (\textbf{Bottom}): the sampled top-1 accuracy curve from the every minimum learning rates in the cycle.}
\label{sec5_fig_lrschedule_imagenet}
\end{figure}
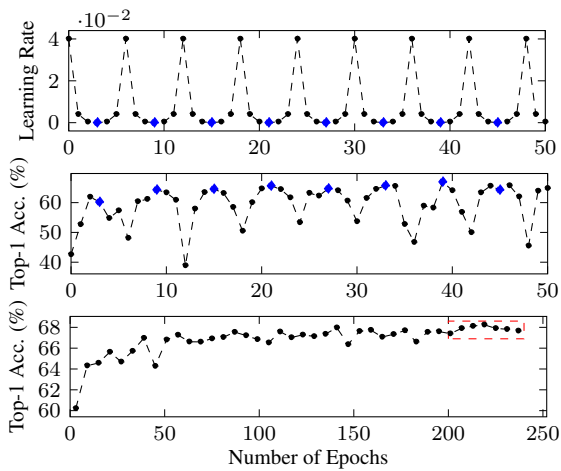

\noindent\textbf{Discussion:} We visualize the SQWA training results using the previous method~\cite{izmailov2018averaging} and our method in~\figurename~\ref{fig_compare_visual_cifar100} (a) and (b), respectively.
The results show similar trends as reported in~~\figurename~\ref{sec5_fig_general} (a) and (b). The original visualization method~\cite{izmailov2018averaging} cannot demonstrate the relationship between the QDNNs. However, our modified method clearly depicts the relationship of the three quantized models. More specifically, we visualized three models from ``the final SQWA model'' ($\mathbf{w3}$), ``the 2-bit quantized version of the averaged model'' ($\mathbf{w1}$), and ``one of the captured models during the cyclical learning rate'' ($\mathbf{w2}$). Thus, $\mathbf{w3}$ can be obtained by fine-tuning $\mathbf{w1}$, and $\mathbf{w2}$ is one of the models to obtain $\mathbf{w1}$. It is noteworthy that all three models were 2-bit ternary QDNNs. \figurename~\ref{fig_compare_visual_cifar100} (a) do not provide a clear correlation of $\mathbf{w1}$, $\mathbf{w2}$, and $\mathbf{w3}$. It shows that the relationship between $\mathbf{w1}$ and $\mathbf{w2}$ is almost similar to that between $\mathbf{w1}$ and $\mathbf{w3}$. 
Our proposed visualization method, as shown in~\figurename~\ref{fig_compare_visual_cifar100} (b), clearly distinguishes the difference between them. $\mathbf{w1}$ is fine-tuned with a low learning rate to obtain $\mathbf{w3}$, and they should exist in the same basin of the loss surface. Furthermore, it is clear that the distance between $\mathbf{w2}$ and $\mathbf{w1}$ is much larger than that between $\mathbf{w3}$ and $\mathbf{w1}$. We expect the proposed visualization method for the QDNNs to be useful for understanding the relationship between quantized networks in future studies.

\subsection{ImageNet}
\label{secsec5_imagenet}
\textbf{Network and hyperparameter configuration:} We trained ResNet-18 \cite{he2016deep} for the ILSVRC 2012 classification dataset~\cite{krizhevsky2012imagenet}. The training hyperparameters are as follows. We trained the full-precision model with a batch size of 1024 on 90 epochs, with the initial learning rate of 0.4 and decreased it by 0.1 times at the 30th, 60th, and 80th epochs. It is noteworthy that the initial learning rate of 0.4 was determined using the \textit{linear scaling rule}, as suggested in~\cite{goyal2017accurate}. We used the SGD optimizer with a momentum of 0.9. Additionally, L2-loss was added with a scale of 1e-4. For the QDNN retraining, the batch size and optimizer were the same as those of the full-precision training. Because we employed SQWA training, cyclical learning rate scheduling was employed, as shown in~\figurename~\ref{sec5_fig_lrschedule_imagenet} (Top). The maximum and minimum values of the learning rates were determined by considering the learning rate of the full-precision training, as suggested in Section~\ref{sec5_SQWA}. 

We captured the models at the minimum points of the cyclical learning rate scheduling and obtained their average. To fine-tune the averaged model, the initial learning rate was set to 0.004 and decreased by 0.1 times at every epoch. We executed five epochs for the fine-tuning and did not employ L2-loss for the QDNN training.

\begin{table}[t]
\centering
\caption{Detailed ImageNet Top-1 and Top-5 accuracies (\%) of the quantized model during retraining with cyclical learning rate scheduling for ResNet18. `Avg.' means the averaged model using seven models that from 202th to 238th epochs, `Direct' represents the direct quantization results of the averaged model, and `Fine-tune' is the result after fine-tuning of direct quantized network.}
\label{table:SQWA_averaged_imagenet}
\begin{tabular}{ccc}
\hline\hline
Epoch (precision) & Top-1 Acc. & Top-5 Acc. \\ \hline
238 (2-bit) & 67.66 & 87.83 \\ 
232 (2-bit) & 67.81 & 88.04 \\ 
226 (2-bit) & 67.90 & 88.00 \\ 
220 (2-bit) & 68.25 & 88.10 \\ 
214 (2-bit) & 68.12 & 88.09 \\ 
208 (2-bit) & 67.90 & 87.89 \\ 
202 (2-bit) & 67.40 & 87.75 \\ \hline
Avg. (4-bit) & 69.66 & 89.12 \\ 
Direct (2-bit) & 60.78 & 83.01 \\ 
Fine-tune (2-bit) & \textbf{69.34} & \textbf{88.77} \\ \hline\hline
\end{tabular}
\end{table}

\noindent\textbf{Results:} Apprentice~\cite{mishra2018apprentice} and QKD~\cite{polino2018model} employed KD to improve the performance of QDNNs. Thus, we employed KD loss for the full-precision training and achieved a top-1 accuracy of 71.68\%. With this full-precision model, we performed SQWA with cyclical learning rate scheduling and captured the quantized models at the lowest leaning rate in the cycles, as described in~\figurename~\ref{sec5_fig_lrschedule_imagenet} (Middle). The accuracy curve of the captured models is depicted in~\figurename~\ref{sec5_fig_lrschedule_imagenet} (Bottom). We adopted the last seven models for averaging and fine-tuning it to obtain the final 2-bit QDNN. The results are reported in~\tablename~\ref{table:SQWA_averaged_imagenet}. The performance of the averaged model is 69.66\%, and we obtained 60.78\% as the direct quantization results. It is noteworthy that the averaged model has a 4-bit precision. After the fine-tuning, the accuracy improved to 69.34\%, which is significantly better than those of the captured models. 

We conducted additional experiments to investigate the effect of number of models on averaging. As discussed in Section~\ref{sec5_SQWA}, the number of captured models is related to the precision of the averaged model. More specifically, the averaged model using three 2-bit ternary models becomes a 3-bit QDNN. Thus, we employed 3, 7, 15, and 31 models such that the precision of the averaged model was 3, 4, 5, and 6 bits, respectively, and fine-tuned each model. The results are reported in~\tablename~\ref{table:SQWA_imagenet_average_precision}. Adopting three models afforded a top-1 accuracy of 69.18\%, which is 0.22\% worse than the seven models. When 15 models were employed, the top-1 accuracy was similar to that of the 7 models but the top-5 accuracy was 0.2\% higher. Using 31 models did not improve the performance. 

We compare our results with those of previous studies in~\tablename~\ref{table:SQWA_compare_imagenet}. Our result outperformed those of previous studies including the 2-bit 4-level (LQ-NET~\cite{zhang2018lq} and WNQ~\cite{cai2019weight}) and ternary (TWN~\cite{fengfu2016ternary}, TTQ~\cite{zhu2016trained}, INQ~\cite{zhou2017incremental}, ADMM~\cite{leng2018extremely}, QNet~\cite{yang2019quantization}, QIL~\cite{jung2019learning}, and Apprentice~\cite{mishra2018apprentice}). More specifically, we achieved a top-1 accuracy of 69.4\%. Only the QNet result is comparable with our result, although it is 0.3\% lower. This result is significant because QNet employs non-linear quantizer while we adopt uniform quantization.

\begin{table}[t]
\centering
\caption{Effect of the number of captured models for averaging. The results are reported in terms of top-1 accuracy after fine-tuning to achive final 2-bit QDNN model on the ImageNet dataset.}
\label{table:SQWA_imagenet_average_precision}
\begin{tabular}{ccccc}
\hline\hline
\begin{tabular}[c]{@{}c@{}}\# of models\\ (bit-precision)\end{tabular} & \begin{tabular}[c]{@{}c@{}}3\\ (3-bit)\end{tabular} & \begin{tabular}[c]{@{}c@{}}7\\ (4-bit)\end{tabular} & \begin{tabular}[c]{@{}c@{}}15\\ (5-bit)\end{tabular} & \begin{tabular}[c]{@{}c@{}}31\\ (6-bit)\end{tabular} \\ \hline
Top-1 Acc. & 69.2 & 69.4 & 69.4 & 69.4 \\ 
Top-5 Acc. & 88.7 & 88.7 & 88.9 & 88.8 \\ \hline\hline
\end{tabular}
\end{table}

\begin{table}[t]
\centering
\caption{Comparison with literature in terms of the validation accuracy (\%) for 2-bit ResNet18 on ImageNet.}
\label{table:SQWA_compare_imagenet}
\begin{tabular}{cccc}
\hline\hline
Methods & Quant Level &Top-1 & Top-5 \\ \hline
TWN~\cite{fengfu2016ternary}& Ternary & 61.8 & 84.2 \\ 
TTQ~\cite{zhu2016trained}&  Ternary & 66.6 & 87.2 \\ 
INQ~\cite{zhou2017incremental}&  Ternary & 66.0 & 87.1 \\ 
ADMM~\cite{leng2018extremely}&  Ternary & 67.0 & 87.5 \\ 
LQ-Net~\cite{zhang2018lq}&  4-level & 68.0 & 88.0 \\ 
QNet~\cite{yang2019quantization}&  Ternary & 69.1 & \textbf{88.9} \\ 
WNQ~\cite{cai2019weight}&  4-level & 67.7 & 87.9 \\ 
QIL~\cite{jung2019learning}&  Ternary & 68.1 & 88.3 \\ 
Apprentice~\cite{mishra2018apprentice}&  Ternary & 68.5 & 88.4 \\ 
SQWA (ours)& Ternary & \textbf{69.4} & \textbf{88.9} \\ \hline
\end{tabular}
\end{table}
\section{Concluding remarks}
\label{sec5_concluding}
We proposed an SQWA algorithm for the optimum quantization of deep neural networks. The model averaging technique was employed to improve the generalization capability of QDNNs by moving them to the wide minimum of the loss surface. Because SQWA captures multiple models for averaging using only a single training with cyclical learning rate scheduling, it is easy to implement and can be applied to many different models. Although we only used a uniform quantization scheme, our results far exceeded the performances of existing non-uniform quantized models in the CIFAR-100 and ImageNet datasets. Additionally, we presented a visualization technique that showed the location of three QDNNs on a single loss surface. Because the proposed method is a training scheme to improve the generalization of QDNNs, it can be combined with other elaborate and non-uniform quantization schemes.

{
\bibliographystyle{ieee_fullname}
\bibliography{egpaper_for_review}

\begin{thebibliography}{10}\itemsep=-1pt

\bibitem{amodei2016deep}
Dario Amodei, Sundaram Ananthanarayanan, Rishita Anubhai, Jingliang Bai, Eric
  Battenberg, Carl Case, Jared Casper, Bryan Catanzaro, Qiang Cheng, Guoliang
  Chen, et~al.
\newblock Deep speech 2: End-to-end speech recognition in english and mandarin.
\newblock In {\em International conference on machine learning}, pages
  173--182, 2016.

\bibitem{ando2017brein}
Kota Ando, Kodai Ueyoshi, Kentaro Orimo, Haruyoshi Yonekawa, Shimpei Sato,
  Hiroki Nakahara, Shinya Takamaeda-Yamazaki, Masayuki Ikebe, Tetsuya Asai,
  Tadahiro Kuroda, et~al.
\newblock Brein memory: A single-chip binary/ternary reconfigurable in-memory
  deep neural network accelerator achieving 1.4 tops at 0.6 w.
\newblock {\em IEEE Journal of Solid-State Circuits}, 53(4):983--994, 2017.

\bibitem{anwar2015fixed}
Sajid Anwar, Kyuyeon Hwang, and Wonyong Sung.
\newblock Fixed point optimization of deep convolutional neural networks for
  object recognition.
\newblock In {\em Acoustics, Speech and Signal Processing (ICASSP), 2015 IEEE
  International Conference on}, pages 1131--1135. IEEE, 2015.

\bibitem{cai2019weight}
Wen-Pu Cai and Wu-Jun Li.
\newblock Weight normalization based quantization for deep neural network
  compression.
\newblock {\em arXiv preprint arXiv:1907.00593}, 2019.

\bibitem{choi2018pact}
Jungwook Choi, Zhuo Wang, Swagath Venkataramani, Pierce I-Jen Chuang,
  Vijayalakshmi Srinivasan, and Kailash Gopalakrishnan.
\newblock {PACT}: Parameterized clipping activation for quantized neural
  networks.
\newblock {\em arXiv preprint arXiv:1805.06085}, 2018.

\bibitem{courbariaux2015binaryconnect}
Matthieu Courbariaux, Yoshua Bengio, and Jean-Pierre David.
\newblock Binaryconnect: Training deep neural networks with binary weights
  during propagations.
\newblock In {\em Advances in Neural Information Processing Systems ({NIPS})},
  pages 3123--3131, 2015.

\bibitem{draxler2018essentially}
Felix Draxler, Kambis Veschgini, Manfred Salmhofer, and Fred~A Hamprecht.
\newblock Essentially no barriers in neural network energy landscape.
\newblock {\em arXiv preprint arXiv:1803.00885}, 2018.

\bibitem{fengfu2016ternary}
Li Fengfu, Zhang Bo, and Liu Bin.
\newblock Ternary weight networks.
\newblock In {\em NIPS Workshop on EMDNN}, volume 118, page 119, 2016.

\bibitem{garipov2018loss}
Timur Garipov, Pavel Izmailov, Dmitrii Podoprikhin, Dmitry~P Vetrov, and
  Andrew~G Wilson.
\newblock Loss surfaces, mode connectivity, and fast ensembling of dnns.
\newblock In {\em Advances in Neural Information Processing Systems}, pages
  8789--8798, 2018.

\bibitem{goyal2017accurate}
Priya Goyal, Piotr Doll{\'a}r, Ross Girshick, Pieter Noordhuis, Lukasz
  Wesolowski, Aapo Kyrola, Andrew Tulloch, Yangqing Jia, and Kaiming He.
\newblock Accurate, large minibatch sgd: Training imagenet in 1 hour.
\newblock {\em arXiv preprint arXiv:1706.02677}, 2017.

\bibitem{guo2017network}
Yiwen Guo, Anbang Yao, Hao Zhao, and Yurong Chen.
\newblock Network sketching: Exploiting binary structure in deep {CNNs}.
\newblock In {\em 2017 IEEE Conference on Computer Vision and Pattern
  Recognition ({CVPR})}, volume~2. IEEE, 2017.

\bibitem{he2016deep}
Kaiming He, Xiangyu Zhang, Shaoqing Ren, and Jian Sun.
\newblock Deep residual learning for image recognition.
\newblock In {\em 2017 IEEE Conference on Computer Vision and Pattern
  Recognition ({CVPR})}, pages 770--778. IEEE, 2016.

\bibitem{hinton2015distilling}
Geoffrey Hinton, Oriol Vinyals, and Jeff Dean.
\newblock Distilling the knowledge in a neural network.
\newblock {\em arXiv preprint arXiv:1503.02531}, 2015.

\bibitem{hochreiter1997flat}
Sepp Hochreiter and J{\"u}rgen Schmidhuber.
\newblock Flat minima.
\newblock {\em Neural Computation}, 9(1):1--42, 1997.

\bibitem{hou2018loss}
Lu Hou and James~T Kwok.
\newblock Loss-aware weight quantization of deep networks.
\newblock {\em arXiv preprint arXiv:1802.08635}, 2018.

\bibitem{hou2016loss}
Lu Hou, Quanming Yao, and James~T Kwok.
\newblock Loss-aware binarization of deep networks.
\newblock {\em arXiv preprint arXiv:1611.01600}, 2016.

\bibitem{hwang2014fixed}
Kyuyeon Hwang and Wonyong Sung.
\newblock Fixed-point feedforward deep neural network design using weights +1,
  0, and -1.
\newblock In {\em Signal Processing Systems (SiPS), 2014 IEEE Workshop on},
  pages 1--6. IEEE, 2014.

\bibitem{izmailov2018averaging}
Pavel Izmailov, Dmitrii Podoprikhin, Timur Garipov, Dmitry Vetrov, and
  Andrew~Gordon Wilson.
\newblock Averaging weights leads to wider optima and better generalization.
\newblock {\em arXiv preprint arXiv:1803.05407}, 2018.

\bibitem{jastrzkebski2017three}
Stanis{\l}aw Jastrzkebski, Zachary Kenton, Devansh Arpit, Nicolas Ballas, Asja
  Fischer, Yoshua Bengio, and Amos Storkey.
\newblock Three factors influencing minima in {SGD}.
\newblock {\em arXiv preprint arXiv:1711.04623}, 2017.

\bibitem{jung2019learning}
Sangil Jung, Changyong Son, Seohyung Lee, Jinwoo Son, Jae-Joon Han, Youngjun
  Kwak, Sung~Ju Hwang, and Changkyu Choi.
\newblock Learning to quantize deep networks by optimizing quantization
  intervals with task loss.
\newblock In {\em Proceedings of the IEEE Conference on Computer Vision and
  Pattern Recognition}, pages 4350--4359, 2019.

\bibitem{krizhevsky2009learning}
Alex Krizhevsky and Geoffrey Hinton.
\newblock Learning multiple layers of features from tiny images.
\newblock Technical report, Citeseer, 2009.

\bibitem{krizhevsky2012imagenet}
Alex Krizhevsky, Ilya Sutskever, and Geoffrey~E Hinton.
\newblock Imagenet classification with deep convolutional neural networks.
\newblock In {\em Advances in Neural Information Processing Systems ({NIPS})},
  pages 1097--1105, 2012.

\bibitem{leng2018extremely}
Cong Leng, Zesheng Dou, Hao Li, Shenghuo Zhu, and Rong Jin.
\newblock Extremely low bit neural network: Squeeze the last bit out with admm.
\newblock In {\em Thirty-Second AAAI Conference on Artificial Intelligence},
  2018.

\bibitem{mishra2018apprentice}
Asit Mishra and Debbie Marr.
\newblock Apprentice: Using knowledge distillation techniques to improve
  low-precision network accuracy.
\newblock In {\em International Conference on Learning Representations}, 2018.

\bibitem{miyashita2016convolutional}
Daisuke Miyashita, Edward~H Lee, and Boris Murmann.
\newblock Convolutional neural networks using logarithmic data representation.
\newblock {\em arXiv preprint arXiv:1603.01025}, 2016.

\bibitem{polino2018model}
Antonio Polino, Razvan Pascanu, and Dan Alistarh.
\newblock Model compression via distillation and quantization.
\newblock In {\em International Conference on Learning Representations}, 2018.

\bibitem{russakovsky2015imagenet}
Olga Russakovsky, Jia Deng, Hao Su, Jonathan Krause, Sanjeev Satheesh, Sean Ma,
  Zhiheng Huang, Andrej Karpathy, Aditya Khosla, Michael Bernstein, et~al.
\newblock Imagenet large scale visual recognition challenge.
\newblock {\em International Journal of Computer Vision}, 115(3):211--252,
  2015.

\bibitem{sandler2018mobilenetv2}
Mark Sandler, Andrew Howard, Menglong Zhu, Andrey Zhmoginov, and Liang-Chieh
  Chen.
\newblock {MobileNetV2}: Inverted residuals and linear bottlenecks.
\newblock In {\em Proceedings of the IEEE Conference on Computer Vision and
  Pattern Recognition}, pages 4510--4520, 2018.

\bibitem{shin2017fixed}
Sungho Shin, Yoonho Boo, and Wonyong Sung.
\newblock Fixed-point optimization of deep neural networks with adaptive step
  size retraining.
\newblock In {\em 2017 IEEE International conference on acoustics, speech and
  signal processing (ICASSP)}, pages 1203--1207. IEEE, 2017.

\bibitem{shin2019empirical}
Sungho Shin, Yoonho Boo, and Wonyong Sung.
\newblock Knowledge distillation for optimization of quantized deep neural
  networks.
\newblock {\em arXiv preprint arXiv:1909.01688}, 2019.

\bibitem{shin2019hlhlp}
Sungho Shin, Jinhwan Park, Yoonho Boo, and Wonyong Sung.
\newblock {HLHL}p: Quantized neural networks training for reaching flat minima
  in loss surface.
\newblock In {\em Thirty-Fourth AAAI Conference on Artificial Intelligence},
  2020.

\bibitem{smith2017cyclical}
Leslie~N Smith.
\newblock Cyclical learning rates for training neural networks.
\newblock In {\em Applications of Computer Vision (WACV), 2017 IEEE Winter
  Conference on}, pages 464--472. IEEE, 2017.

\bibitem{srivastava2014dropout}
Nitish Srivastava, Geoffrey Hinton, Alex Krizhevsky, Ilya Sutskever, and Ruslan
  Salakhutdinov.
\newblock Dropout: a simple way to prevent neural networks from overfitting.
\newblock {\em The Journal of Machine Learning Research}, 15(1):1929--1958,
  2014.

\bibitem{umuroglu2017finn}
Yaman Umuroglu, Nicholas~J Fraser, Giulio Gambardella, Michaela Blott, Philip
  Leong, Magnus Jahre, and Kees Vissers.
\newblock Finn: A framework for fast, scalable binarized neural network
  inference.
\newblock In {\em Proceedings of the 2017 ACM/SIGDA International Symposium on
  Field-Programmable Gate Arrays}, pages 65--74. ACM, 2017.

\bibitem{van2017l2}
Twan van Laarhoven.
\newblock L2 regularization versus batch and weight normalization.
\newblock {\em arXiv preprint arXiv:1706.05350}, 2017.

\bibitem{wan2013regularization}
Li Wan, Matthew Zeiler, Sixin Zhang, Yann Le~Cun, and Rob Fergus.
\newblock Regularization of neural networks using dropconnect.
\newblock In {\em International conference on machine learning}, pages
  1058--1066, 2013.

\bibitem{wu2016google}
Yonghui Wu, Mike Schuster, Zhifeng Chen, Quoc~V Le, Mohammad Norouzi, Wolfgang
  Macherey, Maxim Krikun, Yuan Cao, Qin Gao, Klaus Macherey, et~al.
\newblock Google's neural machine translation system: Bridging the gap between
  human and machine translation.
\newblock {\em arXiv preprint arXiv:1609.08144}, 2016.

\bibitem{yang2019swalp}
Guandao Yang, Tianyi Zhang, Polina Kirichenko, Junwen Bai, Andrew~Gordon
  Wilson, and Christopher De~Sa.
\newblock Swalp: Stochastic weight averaging in low-precision training.
\newblock {\em arXiv preprint arXiv:1904.11943}, 2019.

\bibitem{yang2019quantization}
Jiwei Yang, Xu Shen, Jun Xing, Xinmei Tian, Houqiang Li, Bing Deng, Jianqiang
  Huang, and Xian-sheng Hua.
\newblock Quantization networks.
\newblock In {\em Proceedings of the IEEE Conference on Computer Vision and
  Pattern Recognition}, pages 7308--7316, 2019.

\bibitem{young2018recent}
Tom Young, Devamanyu Hazarika, Soujanya Poria, and Erik Cambria.
\newblock Recent trends in deep learning based natural language processing.
\newblock {\em ieee Computational intelligenCe magazine}, 13(3):55--75, 2018.

\bibitem{zhang2018lq}
Dongqing Zhang, Jiaolong Yang, Dongqiangzi Ye, and Gang Hua.
\newblock Lq-nets: Learned quantization for highly accurate and compact deep
  neural networks.
\newblock In {\em Proceedings of the European Conference on Computer Vision
  ({ECCV})}, pages 365--382, 2018.

\bibitem{zhou2017incremental}
Aojun Zhou, Anbang Yao, Yiwen Guo, Lin Xu, and Yurong Chen.
\newblock Incremental network quantization: Towards lossless cnns with
  low-precision weights.
\newblock {\em arXiv preprint arXiv:1702.03044}, 2017.

\bibitem{zhou2016dorefa}
Shuchang Zhou, Yuxin Wu, Zekun Ni, Xinyu Zhou, He Wen, and Yuheng Zou.
\newblock Dorefa-net: Training low bitwidth convolutional neural networks with
  low bitwidth gradients.
\newblock {\em arXiv preprint arXiv:1606.06160}, 2016.

\bibitem{zhou2017balanced}
Shu-Chang Zhou, Yu-Zhi Wang, He Wen, Qin-Yao He, and Yu-Heng Zou.
\newblock Balanced quantization: An effective and efficient approach to
  quantized neural networks.
\newblock {\em Journal of Computer Science and Technology}, 32(4):667--682,
  2017.

\bibitem{zhu2016trained}
Chenzhuo Zhu, Song Han, Huizi Mao, and William~J Dally.
\newblock Trained ternary quantization.
\newblock {\em International Conference on Learning Representations ({ICLR})},
  2017.

\end{thebibliography}
}

\end{document}